# An Improvement to k-Nearest Neighbor Classifier

*T. Hitendra Sarma, P. Viswanath,
D. Sai Koti Reddy and S. Sri Raghava*

Department of Computer Science and Information Technology
NRI Institute of Technology-Guntur, Guntur

**ABSTRACT**

*K-Nearest neighbor classifier (k-NNC) is simple to use and has little design time like finding k values in k-nearest neighbor classifier, hence these are suitable to work with dynamically varying data-sets. There exists some fundamental improvements over the basic k-NNC, like weighted k-nearest neighbors classifier (where weights to nearest neighbors are given based on linear interpolation), using artificially generated training set called bootstrapped training set, etc. These improvements are orthogonal to space reduction and classification time reduction techniques, hence can be coupled with any of them. The paper proposes another improvement to the basic k-NNC where the weights to nearest neighbors are given based on Gaussian distribution (instead of linear interpolation as done in weighted k-NNC) which is also independent of any space reduction and classification time reduction technique. We formally show that our proposed method is closely related to non-parametric density estimation using a Gaussian kernel. We experimentally demonstrate using various standard data-sets that the proposed method is better than the existing ones in most cases.*

## INTRODUCTION

Nearest neighbor classifier (NNC) and its variants like k- nearest neighbor classifier (k-NNC) are simple to use and often shows good performance [1, 2, 3, 4, 5]. It is shown that when infinite numbers of training patterns are available, k-NNC is equivalent to the Bayes classifier and the error of NNC is less than twice the Bayes error [1]. Performance of k-NNC or NNC with a larger training set is better than that with a smaller training set. It has no significant de- sign (training) phase (except finding k value in case of k-NNC), hence it is highly adoptive to dynamically varying data sets. Hence NNC or k-NNC or its variants are suitable to work with data mining applications where training data sets are large.

Two prominent problems with nearest neighbor based classifiers are (i) the space requirement, which is $O(n)$ where n, is the number of training patterns, and (ii) its classification time is also *O(n)*. That is, one needs to store the entire training set and search it to classify a given pattern. The space requirement problem can be solved to some ex- tent by prototype selection [6, 7, 5], or by reducing the dimensionality by using feature selection and extraction methods [1], or by using a compact representation of the training data like PC-tree [8], FP-tree [9], CF-tree [10], etc., or else by using some editing techniques [3] that reduce the training set size without affecting the performance much. The classification time requirement problem can be solved by finding an index over the training set, like R- Tree [11].

Two fundamental improvements over k-NNC are,(i)weighted k-nearest neighbor Classifier(wk-NNC) [12], where a weight for each training pattern is assigned and is used in the classification, and (ii) generating an artificial training set by applying a bootstrapping method and to use the bootstrapped training set in the place of the original training set. Bootstrap method given by Hamamoto [13] is shown to work well with nearest neighbor based classifiers. Any of these methods can be coupled with any space reduction and or with any of the indexing methods. There exists in Literature another bootstrap method used for k-NNC by Vijaya Saradhi *et al.* [14] where two methods called REST and MREST are given which basically reduces the training set size, but does not improve the performance. Hence this is only a space reduction technique.

The present paper proposes a novel and fundamental improvement over k-NNC called Gaussian weighted k-nearest neighbor classifier (Gwk-NNC). Gwk-NNC assigns a weight to each of the k neighbors based on Gaussian probability density function. It is formally shown that, doing like this is closely related to non-parametric density estimation using a Gaussian kernel. Experimentally using several standard data-sets, it is shown that the proposed method is a better one in most of the cases. Further, the proposed method is also independent of any space reduction and or classification reduction technique, hence can be easily coupled with them.

The rest of the paper is organized as follows. Section 2 describes Notation and definitions used throughout the paper. Section 3 describes about NNC, k-NNC, and Hamamoto's bootstrap method. Section 4 deals with the proposed classifier, *viz.,* the Gaussian weighted k-nearest neighbor classifier (Gwk-NNC). Section 5 gives some of the empirical results and finally Section 6 gives some of the conclusions along with a few future directions of research.

**NOTATIONS AND DEFINITIONS**

**Pattern:** A pattern (data instance) is a $d$ dimensional vector of feature values. For example, $X = (x_1, x_2 \ldots x_d)^T$ is a pattern.

**Set of Features:** The set of features is $F = (f_1, f_2 \ldots f_d)$. $X[f_i]$ is the value of the feature $f_i$ in the instance X. That is, if $X = (x_1, x_2 \ldots x_d)^T$ then $X[f_i] = x_i$. Domain of values for feature $f_l$, for $1 \leq l \leq d$ is $D_l$.

**Feature Space:** It is the set of all patterns denoted by $D = D_1 \times D_2 \times \ldots \ldots D_d$

**Set of Classes:** $\Omega = \{\omega_1, \omega_2, \ldots \omega_k\}$ is the set of classes. We say that the class label of a pattern X is $\omega_i$ to mean the same thing as X belongs to class $\omega_i$.

**Set of Training Patterns (Training Set):**

$\mathcal{X} = \{(X_1, l_1), (X_2, l_2), \ldots, (X_n, l_n)\}$ is the set of all training patterns available by using which the classifier can be found. Here, each of $X_i$, for $1 \leq i \leq n$ is a pattern and $l_i$ is the class to which it belongs. So, $l_i \in \Omega$.

$\mathcal{X}^j$ is the sub-set of training patterns for where for each pattern its class label is $\omega_j$. Here, $1 \leq j \leq c$. Size of $\mathcal{X}^j$ is $n_j$.

$\mathcal{X} = \mathcal{X}^1 \cup \mathcal{X}^2 \cup \ldots \mathcal{X}^c$. Size of $\mathcal{X}$ is $n$. That is, $n = \sum_{j=1}^{c} n_j$

**Set of Test Patterns (Test Set):** This is an independently drawn data-set from the same source as the training set and is denoted by $\mathcal{S}$. For each pattern in $\mathcal{S}$ we know its actual class label.

**Distance Function:** Distance measure used is the Euclidean distance. Between two patterns, $X = (x_1, x_2, \ldots x_d)^T$ and $Y = (y_1, y_2, \ldots, y_d)^T$ is

$$\text{dist}(X, Y) = \sqrt{\sum_{j=1}^{d}(x_j - y_j)^2}$$

**Classifier:** It is a function $f: D \rightarrow \Omega$. This function is found by using the training set.

**Loss Function:** For a pattern Y let its actual class is $\omega_Y$. Then the loss incurred by the classifier for the pattern Y is the function $L_f(Y)$, which is defined below.

$$L_f(Y) = \begin{cases} 0 & \text{if } \omega_Y = f(X) \\ 1 & \text{if } \omega_Y \neq f(X) \end{cases} \quad \ldots (1)$$

This loss function is traditionally called 0-1 loss function. This is used to measure the performance of the classifier over the test set.

**Classification Accuracy (CA):** This is the accuracy of the classifier over the test set. This is normally measured as a percentage. That is, CA is the percentage of patterns in the test set that are correctly classified by the classifier. This is

## NEAREST NEIGHBORS BASED CLASSIFTERS

*The nearest neighbor classifier (NNC)* assigns a class to the given test pattern which is the class of its nearest neighbor in the training set according to the distance function. Let $Y \in y$ be the test pattern. Let $X_N \in x$ be a pattern such that $dist(X_N, Y) \leq dist(X, Y)$ for all X  5. Then we say  is a nearest neighbor.  need not be unique. There can be more than one pattern in the training set which are at equal distance from Y. Let the class of  be. Then the class assigned to the test pattern Y is  In case, if there are several patterns qualifying to be  then a tie occurs. Ties are broken arbitrarily. Let the actual class label of Y which is available with the test set be. Then, if =, we say that the pattern Y is correctly Classified; otherwise we say it is wrongly classified.

*The k-nearest neighbor classifier (k-NNC)* where $k$ is an integer and $k \geq 1$ is a generalization over NNC. $k$ nearest neighbors for the given test pattern Yare found in the training set. The class information of the each of the $k$ nearest neighbors is preserved. Let the $k$ nearest neighbors nearest neighbors along with their class information be $\{(X^1, l^1), (X^2, l^2), \ldots (X^k, l^k)\}$ where the class to which the training pattern $X^1$ belongs is $l^1$, the class to which $X^2$ belongs is $l^2$, and so on. Then the class label assigned to Y is according the majority voting among the labels $l^1, l^2, \ldots, l^k$. That is, for example, assume k = 7, let the class labels of the 7 nearest neighbors be $\omega_2, \omega_1, \omega_2, \omega_2, \omega_2, \omega_3, \omega_3$. Then $\omega_2$ occurring 4 times, where as $\omega_3$ occurred for 2 times and $\omega_1$ occurred for 1 time, the majority vote winner is $\omega_2$. Hence the class label assigned to the test pattern Y is $\omega_2$. If there are more than one majority vote winner then a tie occurs. Ties are broken arbitrarily.

*The weighted k-nearest neighbor classifier (wk-NNC)* which is also known as *the modified k-nearest neighbor classifier* [12] is an improvement over k-NNC. This is achieved by giving weights to each of the k nearest neighbors. Let $w^i = (h^k - h^i)/(h^k - h^1)$ be the weight for the $i^{th}$ nearest neighbor for $1 \leq i \leq k$, where $h^i$ is the distance of $h^i = dist(Y, X^i)$. Total weight for each class is found by summing up the weights of nearest neighbors belonging to that class. If there are no nearest neighbors for a particular class, then the total weight assigned to that class is 0. Now, the test pattern is assigned to the class which has maximum total weight. k-NNC can be seen as a special case of wk-NNC where sum of weights for all *k* nearest neighbors is equal to 1.

Another improvement found in the Literature is to generate an artificial training set from the given training set. Then k-NNC or NNC uses the artificial training set instead of the originally given training set. This process of generating a new training set from the given training set is often called *bootstrapping*. A Bootstrapping method given by Hamamoto *et al.,* [13] generates bootstrap samples (artificial samples) by locally combining original training samples. For a given original training pattern, it first finds *r* nearest neighbors within the class of the given pattern and then finds a weighted average of these neighbors. Let X is a training pattern and let $X^1, X^2, \ldots, X^r$ be its *r* nearest neighbors in its class. Then $X' = \sum_{i=1}^{r} w_i X^i$ (where $\sum_{i=1}^{r} w_i = 1$) is the bootstrapped pattern generated for X. Either all patterns in the original set are considered or else a random sample is taken with replacement for generating bootstrapped patterns. The weights $w_i$ are either chosen to be equal for all *i* (equal to 1/*r*) or else randomly assigned a value (such $\sum_{i=1}^{r} w_i = 1$). So, four different techniques for generating bootstrapped training sets are given. Out of these four methods, it is shown that considering all training patterns by giving equal weight to the *r* neighbors is shown to perform better. So, this method is used for our comparative study. Experimentally it is shown that NNC with bootstrap samples outperform conventional k-NNC in almost all cases.

## GAUSSIAN WEIGHTED K-NEAREST NEIGHBOR CLASSIFIER (GWK-NNC): THE PROPOSED IMPROVEMENT TO K-NNC

This section describes the proposed improvement to k-NNC. The proposed improvement is orthogonal to the space reduction techniques and classification time reduction techniques. Any of the space reduction, as well as classification time reduction techniques can be applied along with the proposed improvement. Similarly this improvement is independent of the bootstrap methods. The improved method is called *Gaussian weighted k-nearest neighbor classifier (Gwk-NNC)* which is experimentally demonstrated to give better classification accuracy when applied with some standard datasets.

The wk-NNC gives weight 1 to the first nearest neighbor and weight 0 to the $k^{th}$ nearest neighbor. The weights of in between neighbors are given based on linear interpolation. See Figure 1.

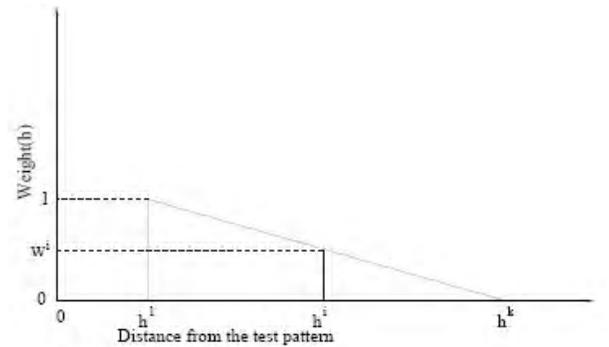

Note: $h^i = dist(Y, X^i)$ = Distance between the test pattern and its i th nearest neighbor

**Fig. 1.** wk-NNC does linear interpolation.

k-NNC is an approximate Bayes classifier [1, 15]. Bayes classifier assigns class label to the test pattern Y based on posterior probabilities, *i.e.,* the class label assigned is $argmax_{\omega_i} = \{P(\omega_1|Y), ..., P(\omega_c|Y)\}$. Here $P(\omega_i|Y)$ is given by

$$P(\omega_i|Y) = \frac{p(Y|\omega_i)P(\omega_i)}{p(Y)}.$$

$p(Y|\omega_i)$, for $1 \leq i \leq c$, is the class conditional probability density of getting pattern Y when given that the class label is $\omega_i$. $P(\omega_i)$ is the prior probability of getting a pattern from class $\omega_i$.

$p(Y|\omega_i)$ can be approximately found by non-parametric methods like Parzen-Windows method [1, 16]. Assume that we draw a hyper-sphere S in the feature space whose volume is V by keeping Y as the centre and which is small enough to encompass the $k$ nearest neighbors of Y. Among the $k$ neighbors let there are $k_i$ neighbors from class $\omega_i$. Then an approximation of $p(Y|\omega_i)$ is $\hat{p}(Y|\omega_i) = \frac{k_i/n_i}{V}$. Here, $n_i$ is the number of training patterns from class. An estimation of $P(\omega_i|Y)$ is

$$\hat{p}(Y|\omega_i) = \frac{\frac{k_i/n_i}{V} \times \frac{n_i}{n}}{p(Y)} = \frac{k_i}{V n p(Y)}.$$

So, $argmax_{\omega_i}\{P(\omega_i|Y)\}_{i=1}^c$ is that class for which $\frac{k_i}{V n p(Y)}$ is maximum. Since the denominator is same for all classes, this is that class for which $k_i$, for $1 \leq i \leq c$, is maximum. This is what the k-NNC is doing and hence is an approximate Bayes classifier. This is the reason for calling k-NNC as a non-parametric classifier. It intrinsically estimates probability densities non-parametrically.

In k-NNC, class conditional densities are found approximately. When a pattern falls in the hyper-sphere it is counted, otherwise when it falls outside, it is not counted. It can be shown that $\hat{p}(Y|\omega_i) \to p(Y|\omega_i)$ as $n \to \infty$, $k \to \infty$, $V \to 0$, and $k/n \to 0$. An improvement over this type of density estimation is to use a kernel function whose general form is

$$\hat{p}(Y|\omega_i) = \frac{1}{n_i} \sum_{\forall X \in x_i} \delta(Y-X)$$

Where $\delta(.)$ is a kernel function with localized support and its exact form depends on $n_i$.

Often Gaussian kernel functions are chosen for two reasons. First, the Gaussian function is smooth and hence the estimated density function also varies smoothly. Second, if we assume a special form of the Gaussian family in which the function is radially symmetrical, the function can be completely specified by a variance parameter only [17]. Thus $\hat{p}(Y|\omega_i)$ can be expressed as a mixture of radially symmetrical Gaussian kernels with common variance $\sigma^2$:

$$\hat{p}(Y|\omega_i) = \frac{1}{n_i (2\pi)^{d/2} \sigma^d} \sum_{\forall X \in x_i} exp\left\{-\frac{\|Y-X\|^2}{2\sigma^2}\right\} \qquad \ldots (2)$$

where $d$ is the dimensionality of the feature space.

Equation (2) can be used to find the class conditional densities and can be used to find posterior probabilities. But the difficulty is, it uses all patterns from a class. An index structure like R-tree can be used to select the first few nearest neighbors efficiently, with a very small amount of time. But this is useless when we are considering all training patterns. Hence, we propose to take into account only patterns in the $k$ nearest neighbors of Y in calculating the class-conditional density.

In $k$ nearest neighbors of Y, let there are $k_i$ patterns from class $\omega_i$, where $1 \leq i \leq c$.

Let these $k_i$ nearest neighbors be $\{X^{(1,i)}, X^{(2,i)}, \ldots X^{(k_i,i)}\}$. Then the class conditional density is

$$\hat{p}(Y|\omega_i) = \frac{1}{k_i (2\pi)^{d/2} \sigma^d} \sum_{j=1}^{k_i} exp\left\{-\frac{\|Y-X^{(j,i)}\|^2}{2\sigma^2}\right\} \qquad \ldots (3)$$

We assume that the data-set is zero mean and unit variance normalized. This is done for both training and test sets, together. Hence, we take ó = 1, i.e., unit variance, So,

$$\hat{p}(Y|\omega_i) = \frac{1}{k_i(2\pi)^{d/2}} \sum_{j=1}^{k_i} exp\left\{-\frac{\|Y-X^{(j,i)}\|^2}{2}\right\} \qquad \ldots (4)$$

Since, we took into account only *k* nearest neighbors, and out of this $k_i$ are from class $\omega_i$, it is more appropriate to say that the apriori probability for class $\omega_i$ is P $(\omega_i) = k_i/k$. Then the posterior probability is

$$\hat{p}(\omega_i|Y) = \frac{1}{k_i(2\pi)^{d/2}} \sum_{j=1}^{k_i} exp\left\{-\frac{\|Y-X^{(j,i)}\|^2}{2}\right\} \frac{(k_i/k)}{p(Y)} \qquad \ldots (5)$$

Now the approximate Bayes classification is to assign the class for which the posterior probability is maximum. Eliminating the terms from Equation 5 which are same for each class, we need to find the class for which

$$W_i = \sum_{j=1}^{k_i} exp\left\{-\frac{\|Y-X^{(j,i)}\|^2}{2}\right\} \qquad \ldots (6)$$

is maximum.

$W_i$ can be seen as the cumulative weight for class $\omega_i$, i.e., $W_i = \sum_{j=1}^{k_i} \omega^{(j,i)}$ where

$$\omega^{(j,i)} = exp\left\{-\frac{\|Y-X^{(j,i)}\|^2}{2}\right\}$$

which is the weight assigned to the neighbor (one among the *k* nearest neighbors) $X^{(j,i)}$ which belongs to class $\omega_i$.

To simplify the notation, assuming the *k* nearest neighbors of Y are $X^1, X^2, \ldots, X^k$, weight assigned to the neighbor $X^i$ is

$$\omega^i = exp\left\{-\frac{(h^i)^2}{2}\right\} \qquad \ldots (7)$$

where $h^i = dist(Y, X^i) = \|Y - X^i\|$.

The function showing weights against distances from the test pattern is shown in Figure 2.

The proposed Gaussian weighted k-nearest neighbor classifier (Gwk-NNC) is similar to the weighted k-NNC, except that the weight for each nearest neighbor (in the *k* nearest neighbors) is given according to Equation 7. The classifier assigns class label

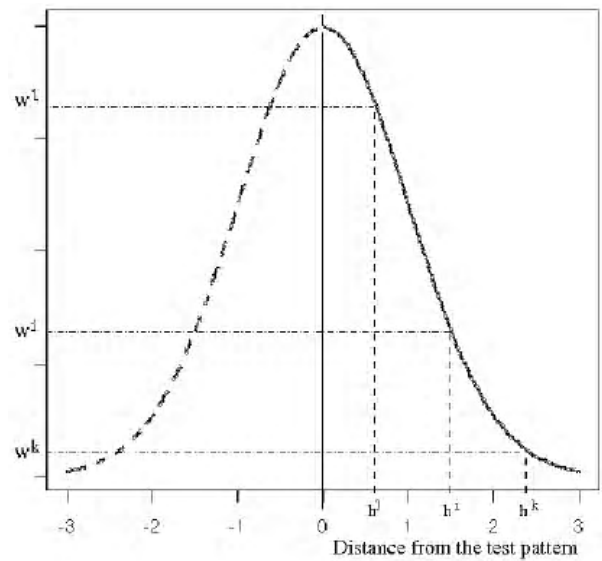

**Fig. 2.** Gaussian weights to nearest neighbors

according to the maximum cumulative weight per class

## EXPERIMENTAL STUDY

This section describes the experimental studies performed.

We performed experiments with five different data-sets, *viz.,* OCR, WINE THYROID, and PENDIGITS, respectively. Except the OCR data-set, all others are from the UCI Repository [18]. OCR dataset is also used in [19, 8]. The properties of the data-sets are given in Table 1. For OCR, THYROID and PENDIGITS data-sets, the training and test sets are separately available. For the WINE dataset 100 patterns are chosen randomly as the training patterns and the remaining as the test patterns. All the data-sets have only numeric valued features. All data-sets are normalized to have zero mean and unit variance for each feature.

Table 2 compares classification accuracies obtained for various classifiers. Classifiers compared are NNC, k-NNC, wk-NNC, k-NNC using Hamamoto's bootstrapping (k-NNC(HBS)) and the proposed classifier Gaussian weighted k-NNC (Gwk-NNC). The parameters like $k$ value and $r$ value (used in Hamamoto's bootstrapping) are found by employing a 3-fold cross validation [1]. To obtain, standard deviation for the measured CA, experiments are done with 10 different test sets which are generated by employing sampling with replacement with the given test set. Table 2 shows the average CA along with standard deviation obtained.

From the presented results, it is observed that the proposed Gwk-NNC performs better than the other related classifiers, in most of the cases.

## CONCLUSIONS AND FUTURE WORK

The paper presented a novel improvement to the conventional k-NNC where the weights to the *k* nearest neighbors is found based on Gaussian distribution, in contrast to wk-NNC where weights are given to the neighbors based on a linear function. The proposed method, Gwk-NNC performs better than the other related classifiers like NNC, k-NNC, wk-NNC and k-NNC using Hamamoto's bootstrap method.

Future directions of research are, (i) to show theoretically the reasons behind good performance of Gwk-NNC, and (ii) to carry out a bias variance analysis for the proposed method which clearly shows which part of the error, *viz.,* bias or variance or both that is reduced by Gwk-NNC when compared with other related classifiers.

**Table 1**: Properties of the data-sets used

| Dataset | Number of features | Number of classes | Number of training examples | Number of test examples |
|---|---|---|---|---|
| OCR | 192 | 10 | 6670 | 3333 |
| WINE | 13 | 3 | 100 | 78 |
| THYROID | 21 | 3 | 3772 | 3428 |
| PENDIGITS | 16 | 10 | 7494 | 3498 |

**Table 2**: A comparison between the classifiers (showing CA $\pm$ std.dev. (in %))

| Dataset | NNC | k-NNC | wk-NNC | k-NNC(HBS) | Gwk-NNC |
|---|---|---|---|---|---|
| OCR | 90.75 $\pm$ 0.51 | 90.93 $\pm$ 0.50 | 91.23 $\pm$ 0.49 | 91.29 $\pm$ 0.48 | 92.01 $\pm$ 0.47 |
| WINE | 94.87 $\pm$ 2.50 | 96.15 $\pm$ 2.18 | 97.44 $\pm$ 1.79 | 97.44 $\pm$ 1.81 | 97.44 $\pm$ 1.80 |
| THYROID | 93.14 $\pm$ 0.43 | 94.40 $\pm$ 0.39 | 94.81 $\pm$ 0.37 | 94.54 $\pm$ 0.38 | 95.07 $\pm$ 0.34 |
| PENDIGITS | 96.08 $\pm$ 0.33 | 97.54 $\pm$ 0.31 | 97.63 $\pm$ 0.30 | 97.54 $\pm$ 0.32 | 97.63 $\pm$ 0.29 |